\documentclass[10pt,twocolumn,letterpaper]{article}

\usepackage{cvpr}
\usepackage{times}
\usepackage{epsfig}
\usepackage{graphicx}
\usepackage{amsmath}
\usepackage{amssymb}
\usepackage{url}
\usepackage{listings}
\usepackage{color}
\usepackage{verbatim}
\usepackage{multirow}
\usepackage{epsfig}
\usepackage{subfigure}
\usepackage{gensymb}
\usepackage{multirow}
\usepackage[ruled, lined,boxed,commentsnumbered]{algorithm2e}

\makeatletter
\newcommand{\removelatexerror}{\let\@latex@error\@gobble}
\makeatother

\def\ie{\emph{i.e}\onedot} 
\def\vec#1{\mbox{\boldmath $#1$}}

\usepackage{soul}

\newcommand{\weichiu}[1]{}

\usepackage[pagebackref=true,breaklinks=true,letterpaper=true,colorlinks,bookmarks=false]{hyperref}

\cvprfinalcopy 

\def\cvprPaperID{249} 

\ifcvprfinal\fi
\begin{document}

\title{Forecasting Interactive Dynamics of Pedestrians with Fictitious Play}

\author{Wei-Chiu Ma$^{1}$ \; De-An Huang$^2$ \; Namhoon Lee$^3$\; Kris M. Kitani$^4$ \\ \\
$^1$MIT \quad $^2$Stanford \quad $^3$Oxford \quad $^4$CMU
}

\maketitle
\begin{abstract} 
We develop predictive models of pedestrian dynamics by encoding the coupled nature of multi-pedestrian interaction using game theory and deep learning-based visual analysis to estimate person-specific behavior parameters. We focus on predictive models since they are important for developing interactive autonomous systems (\eg, autonomous cars, home robots, smart homes) that can understand different human behavior and pre-emptively respond to future human actions. Building predictive models for multi-pedestrian interactions however, is very challenging due to two reasons: (1) the dynamics of interaction are complex interdependent processes, where the decision of one person can affect others; and (2) dynamics are variable, where each person may behave differently (\eg, an older person may walk slowly while the younger person may walk faster). We address these challenges by utilizing concepts from game theory to model the intertwined decision making process of multiple pedestrians and use visual classifiers to learn a mapping from pedestrian appearance to behavior parameters. We evaluate our proposed model on several public multiple pedestrian interaction video datasets. Results show that our strategic planning model predicts and explains human interactions $25\%$ better when compared to a state-of-the-art activity forecasting method.
\end{abstract}

\section{Introduction}

\begin{figure}[t]
  \centering
    \includegraphics[width=1.0\linewidth]{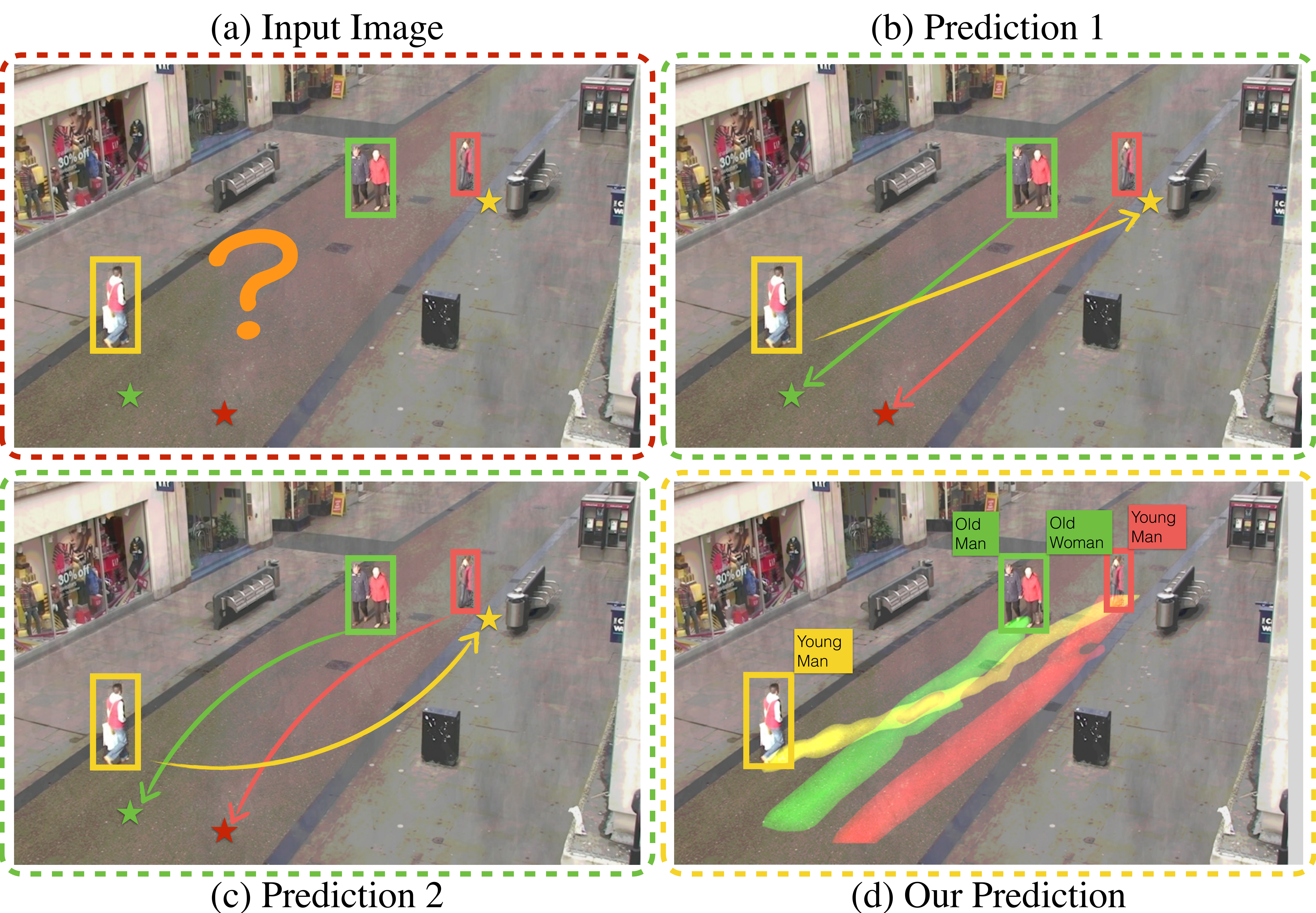}
\caption{Can you forecast their future behavior? A single image contains rich information about future trajectories.}
\label{fig:teaser}
\vspace{-5mm}
\end{figure}

The goal of this work is to imitate the predictive abilities of human cognition, by building a predictive model that takes into account complex reasoning about: (1) the interdependent interactions of multiple pedestrians and (2) important visual cues needed to infer individual behavior patterns. Consider the complexities of predicting the trajectories of multiple pedestrians from a \textit{single} image, as depicted in Figure~\ref{fig:teaser} where four pedestrians are walking on the street. Given this single image, what would one forecast as their future trajectories? A simple prediction would be that all people will walk in a straight line (\ie, the minimum distance) to their goal as in Figure~\ref{fig:teaser}(b). This strategy, however, might lead to collisions between pedestrians (\eg, the young man (yellow) and the older couple (green) may collide). A more thoughtful model might consider the possibility that one or more pedestrians will alter their trajectory based on their \textit{prediction of other pedestrians} (Figure~\ref{fig:teaser}(c)). 
Going further, a more informed model might attempt to take into account the observation that on average, elderly couples tend to walk at a slower rate, while a young man is more likely walk quickly and take pre-emptive maneuvers (Figure~\ref{fig:teaser}(d)). 
Taking these observations into consideration, we might hypothesize that the younger man is more likely to exemplify preemptive avoidance behaviors and weave through the two pedestrians.
This illustration serves to highlight the complex reasoning that is involved in predicting the walking trajectories of several people given limited amount of information (in our scenario a single image). Our goal is to mimic -- computationally -- this ability to reason about the dynamics of interactive social processes. 

Developing computational models for interactive dynamics among humans, however, is an extremely challenging task. This requires a deep understanding of the complex and often subtle norms of human interactions. Pioneering works have attempted to address this by parameterizing human behaviors with models such as social forces \cite{helbing1995social,pellegrini2009you}, potential fields \cite{alahi2014socially}, and flow fields \cite{ali2007lagrangian,ali2008floor}. Yet most of these works are performing either long-term prediction in static environments or short-term prediction in dynamic environments. They do not address the interactions that could occur in the distant future, and do not resolve the long-term prediction problem in dynamic environments. To address these complexities of multi-agent future prediction, we propose a game-theoretic approach. 




We directly address the interdependent nature of human interactions using the language and concepts of multi-player game theory. In particular, we utilize Brown's \cite{brown1951iterative} classical notion of \textit{Fictitious Play} to model the interaction between multiple pedestrians. Brown's fictitious play model assumes that each player will take the best next action based on an observed empirical distribution over the past strategies of other players. As we will show, the multi-player game model has strong parallels to multi-pedestrian forecasting, as each pedestrian pre-emptively plans her path according to beliefs about how other pedestrians will move in the future.

\begin{figure}[bt]
  \centering
    \includegraphics[width=0.96\linewidth]{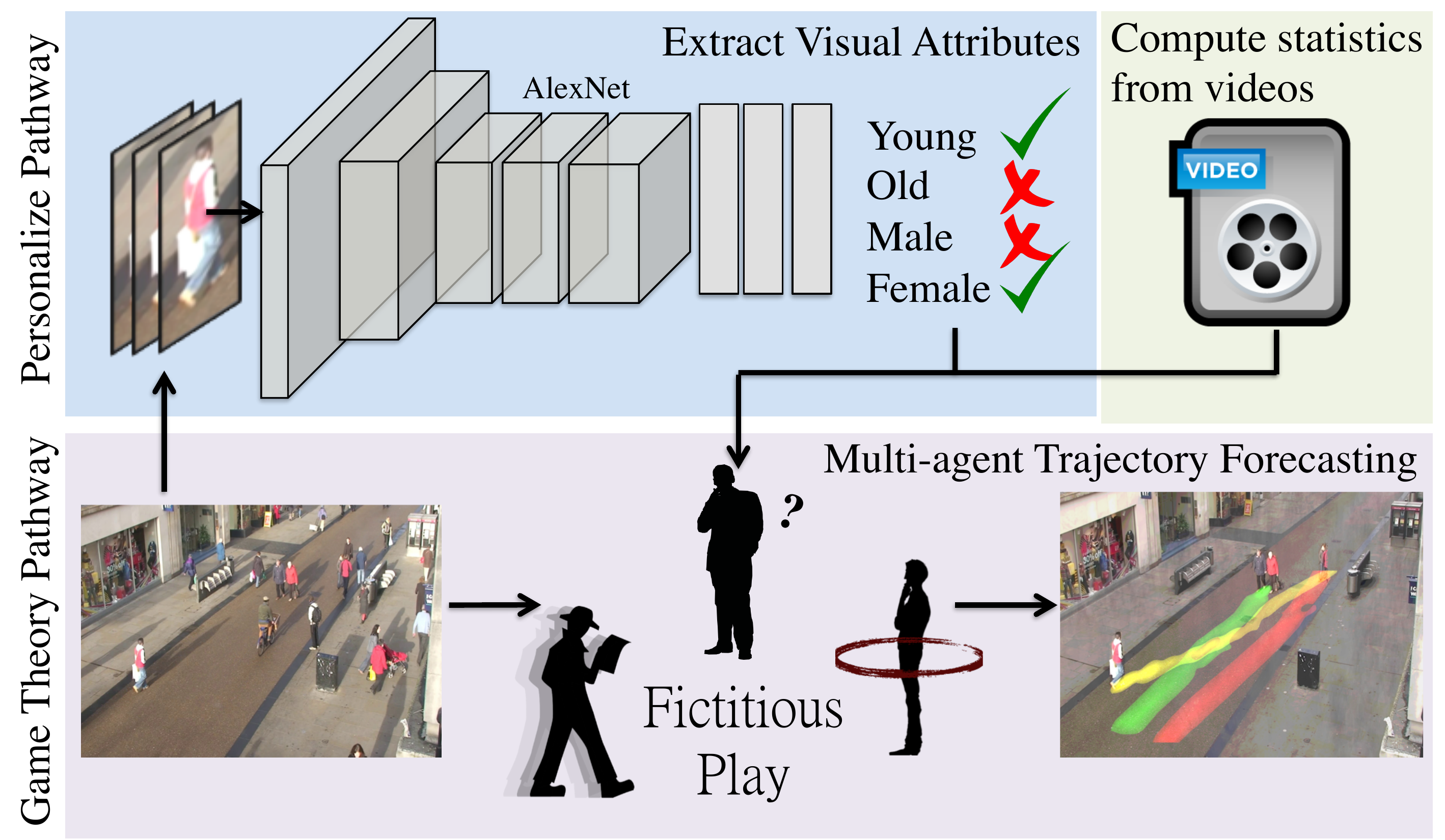}
\caption{Model Overview. \textbf{Personalization pathway} (\textbf{top}) estimates physical properties for each pedestrian based on visual information and statistics from videos. \textbf{Game theory pathway} (\textbf{bottom}) takes as input: (1) estimated properties, (2) individualized motion model per pedestrian. As output, it forecasts multi-pedestrian interactions/trajectories using Fictitious Play.}
\vspace{-5mm}
\label{fig:system_flow}
\end{figure}

To individualize the pedestrian model, we train a deep learning-based classifier to learn visual cues that are indicative of behavior patterns (\eg, age can affect speed). We use the classifier to estimate each pedestrian's velocity based on sub-population statistics. Furthermore, we visually estimate the initial body orientation such that the model is more likely to predict motion aligned to body direction at the start of a predicted trajectory. In this way, we integrate visual analysis with our prediction model. Figure~\ref{fig:system_flow} shows the overview of our approach.

\noindent\textbf{Contributions:} We present a novel technique to forecast multi-pedestrian trajectories from a \textit{single} image. First, we explicitly model the interplay among multiple people by drawing connections between game theory and optimal control. To the best of our knowledge, Fictitious Play has never been applied in the context of modeling pedestrian motions. Second, we address the variability among people by building individualized predictive pedestrian models. We are the first attempt to infer physical properties of each pedestrian from appearance for multi-agent forecasting.

\section{Related Work}

There has been growing interest in developing computational models of human activities that can extrapolate unseen information and predict future unobserved activities \cite{scovanner2009learning,tastan2011leveraging,kretzschmar2014learning,ryoo2015robot,karasevAHS16,walker2015dense,vondrick2015anticipating,soran2015generating,huang2014action,huang2015approximate,jain2015car,jain2015recurrent,rcrf-sener-saxena-rss2015,cancela2014unsupervised,zunino2016intention}. In the context of pedestrian dynamics, Helbing and Molonar \cite{helbing1995social} first integrated the concept of the \textit{social force} model into a computational framework for understanding pedestrian dynamics. Their work incorporated ideas of goals, desired speed and the repulsion due to territorial affects of social forces. In computer vision, the social force model has been used to help aid visual tracking \cite{pellegrini2009you} and anomaly detection \cite{mehran2009abnormal}. More recent work has focused on discovering the underlying potential field by observing human behavior such as patterns of motions \cite{alahi2014socially}, mutual gaze or regions of repulsion. In high-density crowds, the patterns of motion of people can be used to infer an underlying flow field for a given scene \cite{ali2007lagrangian,ali2008floor} and the interaction  between stationary crowds and pedestrians can be used to predict pedestrians' future motions \cite{yi2015understanding}. The global motion or the joint attention of sparse groups of people (e.g., sports scenarios) can also be used to infer basins of attractions or socially salient hot spots \cite{kim2010motion,park20123d}. Patterns of  avoidance can also be used to learn the hidden rewards or costs of physical spaces \cite{kitani2012activity,xie2013inferring,walker2014patch}. 

To make reliable predictions about the long-term future, many techniques often assume a static environment \cite{kitani2012activity,walker2014patch}. In a static environment, the cost topology is constant, where the environment and features do not change over time. In dynamic environments, the cost topology of the state space is constantly changing which means that any computational model must be continually updated. When the cost topology can be accurately updated over time, it can be used for short-term prediction \cite{pellegrini2009you,gong2011multi,kim2014brvo} (or at least until the next update). As such, these techniques have been very effective for tracking multi-pedestrian trajectories. While methods have been proposed for long-term prediction in static environments and short-term prediction in dynamic environments, the task of long-term prediction in dynamics environments remains relatively unexplored in human activity analysis except~\cite{alahi2016social,leepredicting,karasevAHS16}. Concurrent with our work, \cite{alahi2016social} introduced a data-driven approach to \emph{implicitly} encode the interactive dynamics among people. Their model, however, focused only on trajectory data. They ignored the rich information underlying the visual data. 
In \cite{leepredicting,karasevAHS16}, the complex and intertwined interactions between agents is either ignored~\cite{karasevAHS16} or restricted to the perspective of a single agent (only the wide receiver in ~\cite{leepredicting}). In constrast, we directly address the interdependent nature of human interactions using Fictitious Play and perform long-term prediction for \emph{all} of the agents in the scene.



\begin{figure*}[tb]
\begin{center}
\includegraphics[width=1\linewidth, trim = 0 25 0 0]{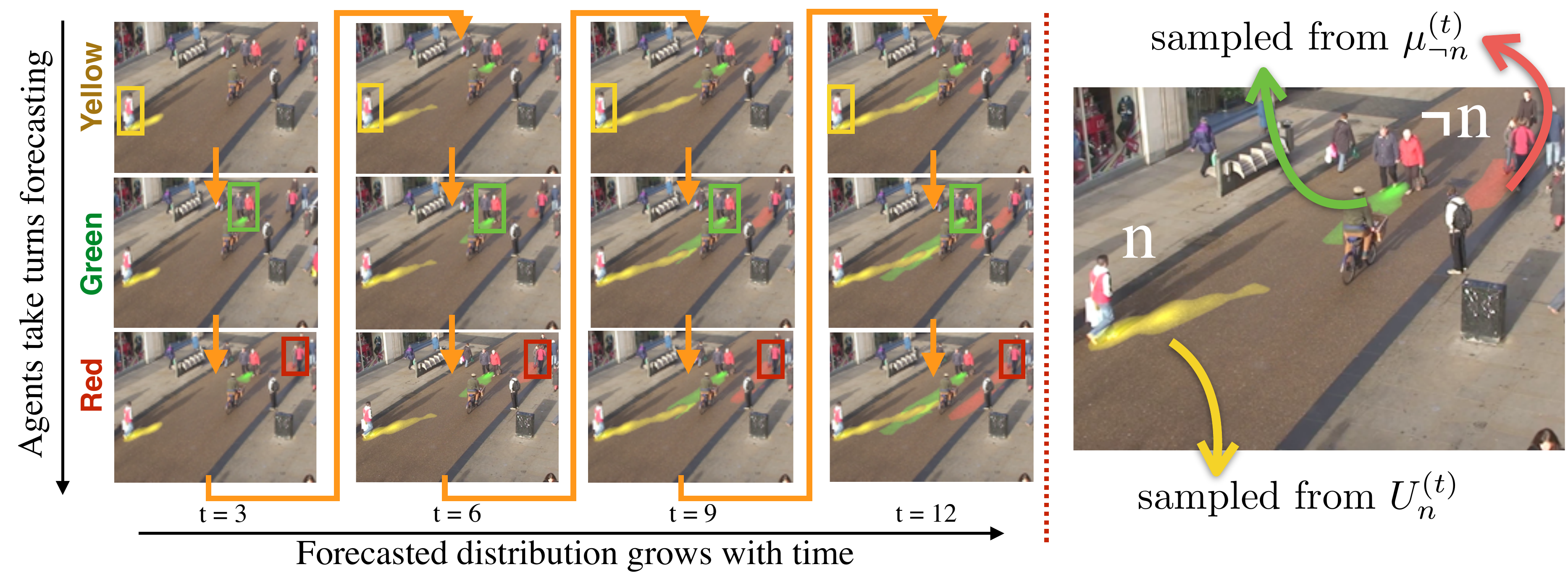}
\end{center}
\caption{(Left) Visualization of Fictitious Play with three pedestrians. (Right) Distributions over states sampled from $U_{n}^{(t)}$ and $\mu_{\neg n}^{(t)}$.}
\label{fig:fp-flow}
\vspace{1mm}
\end{figure*}

\section{Forecasting Multi-Pedestrian Trajectories}

Given a single image and initial pedestrians detections, we aim to develop a predictive model that can forecast plausible future trajectories for all pedestrians. To do this, we must model the complex predictive interplay between multiple pedestrians, while also considering individual differences that might impact behavior, to obtain accurate predictions. To address these challenges, we utilize concepts from game theory to model the intricately coupled interactive prediction process. We also leverage recent success of deep neural networks to infer individual behavior models for each pedestrian from visual evidence. We describe how game theory can be used to frame our multi-pedestrian forecasting problem in Section \ref{sec:fictitious} and present a method for mapping the visual appearance of pedestrians to estimate person-specific behavior parameters in Section \ref{sec:appearance}.

\noindent\textbf{Notation}. We will define the state space (the ground plane) as a 2D lattice, where each position is denoted by $\vec{x} =[x,y] \in \vec{X}$. A pedestrian can make a transition from state to state by taking an action $\vec{a} \in A$ which in the case of a 2D lattice (grid world) is the velocity $[ \dot{x}, \dot{y}]$. A trajectory is a sequence of state-action pairs, $\vec{s}=\{(\vec{x}_1,\vec{a}_1), \ldots, (\vec{x}_K,\vec{a}_K)\}$. Each state $\vec{x}$ has an associated vector of features $\vec{f}(\vec{x}) = [ f_1(\vec{x}) \ldots f_J(\vec{x})]$, where $f_j(\vec{x})$ represent properties of that state such as the output of a visual classifier, the distance to an object or predicted presence of another pedestrian.

\subsection{Forecasting Interactions as Fictitious Play}
\label{sec:fictitious}

Game theory \cite{myerson2013game} is a widely applicable discipline that aims to model adversarial and collaborative interactions between \textit{rational} decision-makers. It has been applied to a range of disciplines including economic theory \cite{rabin1993incorporating}, politics \cite{brams2011game} and computer science \cite{nisan2007algorithmic}. More importantly, it is well-suited for modeling our multi-pedestrian prediction scenario, as the social dynamics of collision avoidance can be modeled as a collaborative multi-player game. To forecast long-term trajectories of multiple pedestrians, we utilize Fictitious Play (FP) \cite{brown1951iterative}, where we model each pedestrian to take a path based on her own predictions of how other pedestrians will move. By incrementally forward simulating pedestrian paths with this model, we can obtain a distribution over possible future paths over multiple people. 

Formally, each pedestrian $n\in \{1,\ldots,N\}$ has the ability to choose a macro-action $\vec{s}_n \in S$ from a set of macro-actions. In our scenario, a macro-action $\vec{s}_n$ is a very short trajectory whose length $L_n$ depends on the speed of the pedestrian $n$ (detailed in Section \ref{sec:appearance}). Each pedestrian has an utility function $U_{n}[\vec{s}_{n}, \mu_{\neg n}(\vec{s}_{\neg n})]$ that maps a given macro-action to a value $U_n:\vec{s}_n \rightarrow \mathbb{R}$. Intuitively, the utility function $U_{n}$ describes the reward of taking a certain path. If there is a low potential of collision, its utility will be high. Notice that the $U_{n}$ is also dependent on the forecasted distributions over macro-actions of all other pedestrians $\mu_{\neg n} (\vec{s}_{\neg n})$. This is needed to compute the potential of collision with other pedestrians. The set of trajectories $\vec{s}_{\neg n}$ is a set of macro-actions of all other pedestrians, $\vec{s}_{\neg n} = \{ \vec{s}_m | m\neq n \}$. We visualize a distribution over states sampled from $\mu_{\neg n}$ in Figure \ref{fig:fp-flow} (right). 

\begin{algorithm}[tb]
\footnotesize   
\DontPrintSemicolon
\SetKwFunction{Union}{Union}\SetKwFunction{FindCompress}{FindCompress}
\SetKwInOut{Input}{Input}\SetKwInOut{Output}{Output}
\Input{Initial state $\vec{x}_{0,n} \forall n$, $\tau$}
\Output{Forecasted cumulative state visitation distribution $\{\bar{D}_n\}$ }
        $D_n^{(0)}(\vec{x}_{0,n}) = 1$ for all $n$\;
        \For{$t = \tau:\tau:T$}{
            \For{$n = 1:N$} {
                $\mu_{\neg n}^{(t)} ~~\leftarrow 
                \textsc{UpdateEmpirical}(\{\mu_m^{(t-\tau)}|m \neq n\})$\hspace{17mm} (Eq.\ref{eq:utility-1}) \;
                $f_{n,soc}^{(t)} \leftarrow 
                \textsc{EncodeToFeature}(\mu_{\neg n}^{(t)}, D^{(t-\tau)}_{\neg n})$
                \hspace{8mm} (Alg.\ref{alg:encode-feature})\;
                $U_n^{(t)} ~~\leftarrow \textsc{UpdateUtility}(f^{(t)}_{n,soc})$ 
                \hspace{30mm} (Eq.\ref{eq:utility-2})\;
                $D_n^{(t:t+L_n)} \leftarrow  
                \textsc{TakeMacroAction} (U_n^{(t)},D_n^{(t-\tau)})$
                \hspace{11mm} (Alg.\ref{alg:macro-actions})\;
                $\bar{D}_n = \bar{D}_n + \sum_{l=t}^{t+\tau-1} D_n^{(l)}$\;
            }
        }
\caption{Multi-Pedestrian Fictitious Play}
\label{alg:fp-mdp}
\end{algorithm}

Algorithm \ref{alg:fp-mdp} describes the process of Fictitious Play. For every forecasting period $\tau$, each pedestrian $n$ forms beliefs about the future actions of other pedestrians by updating the empirical distribution $\mu^{(t)}_{\neg n}$ using the function \textsc{UpdateEmpirical}. Then the distribution $\mu^{(t)}_{\neg n}$ is encoded as social feature $f_{n,soc}^{(t)}$ using the function \textsc{EncodeToFeature}. The utility function of the $n$-th pedestrians is updated according to this new feature with the function \textsc{UpdateUtility}. In the final step, we forecast the movement of the pedestrian with the function \textsc{TakeMacroAction}. This process is repeated for $T$ time steps.

\begin{figure*}[bt]
\vspace{-5mm}
 \begin{minipage}[t]{0.48\linewidth}
 \begingroup
\removelatexerror
\begin{algorithm}[H]
\footnotesize
\caption{\textsc{\footnotesize EncodeToFeature}}
\label{alg:encode-feature}
\DontPrintSemicolon
\SetKwFunction{Union}{Union}\SetKwFunction{FindCompress}{FindCompress}
\SetKwInOut{Input}{Input}\SetKwInOut{Output}{Output}
\Input{Empirical distribution
$\mu_{\neg n}^{(t)}$,
State visitation distribution $D_{\neg n}^{(t-1)}$
}
\Output{Feature vector $f^{(t)}_{n,soc}$}
$f^{(t)}_{n,soc} = \vec{0}$\;
\For{$m = 1:N\ \textrm{and}\ m \neq n$}
{
    $D^{(t:t+L_m)}_{m} \leftarrow \textsc{TakeMacroAction}(\mu^{(t)}_m, D_{m}^{(t-1)})$\;
    $\bar{D}_m = \sum_{l=t}^{t+L_m} D_m^{(l)}$\;
    $f^{(t)}_{n,soc} = f^{(t)}_{n,soc} + \bar{D}_{m}$\;
}
\end{algorithm}
\endgroup
 \end{minipage}
 \begin{minipage}[t]{0.48\linewidth}
  \begingroup
\removelatexerror
\begin{algorithm}[H]
\footnotesize
\caption{\textsc{\footnotesize TakeMacroAction}}
\label{alg:macro-actions}
\DontPrintSemicolon
\SetKwFunction{Union}{Union}\SetKwFunction{FindCompress}{FindCompress}
\SetKwInOut{Input}{Input}\SetKwInOut{Output}{Output}
\Input
{
Empirical distribution $\mu_{n}^{(t)}$,
Prior state visitation distribution $D_{n}^{(t-1)}$
}
\Output
{Future state visitation distributions $D_n^{(t:t+L_n)}$}
\vspace{2mm}
    $\pi(\vec{a}| \vec{x}) \leftarrow  \textsc{ComputePolicy}(\mu_n)$\;
    \For{$l = t:t+L_n$}
    {
        $D_n^{(l)}(\vec{x}') = \sum_{\vec{a},\vec{x}} 
        P(\vec{x}'|\vec{x},\vec{a}) \times $\\
        \hspace{20mm}$\pi(\vec{a}| \vec{x})D^{(l-1)}(\vec{x})$\;
        \hspace{35mm}$\forall \vec{x}'$\;
    }
\end{algorithm}
\endgroup
 \end{minipage}
\vspace{-5mm}
\end{figure*}

\noindent\textbf{\textsc{UpdateEmpirical}}. Under the assumptions of fictitious play, the empirical distribution over opponent macro-actions $\mu_{\neg n} (\vec{s}_{\neg n})$ is typically computed by counting how many times each macro-action was chosen by each player. In our case, we need to describe a distribution over trajectories and so we use a parameterized form of the empirical distribution (\ie, a maximum entropy distribution). The empirical distribution over macro-actions of all other pedestrians is decomposed into a product of distributions for each pedestrian $\mu_{\neg n} (\vec{s}_{\neg n}) \propto \prod_{m\neq n} \mu_m (\vec{s}_m)$. Each distribution is parametrized by a maximum entropy probability (also called Boltzmann or Gibbs) distribution,
\begin{align}
\mu_m (\vec{s}_m) \propto \exp \sum_{\vec{x} \in \vec{s}_m} \vec{\theta}^{\top}\vec{f}_m(\vec{x}),
\label{eq:utility-1}
\end{align}
where $\vec{f}_m(\vec{x})$ are the features of a state $\vec{x}$ along the trajectory $\vec{s}_m$ for the pedestrian $m$, which are weighted by the vector of parameters $\vec{\theta}$. We will explain in Section \ref{sec:ioc} how the parameters $\vec{\theta}$ of the empirical distribution are learned from a dataset of demonstrated pedestrian behavior.

\noindent\textbf{\textsc{EncodeToFeature}}. This function maps $\mu^{(t)}_{\neg n}$ to the feature vector $f_{n,soc}^{(t)}$. Intuitively, this function predicts how all other pedestrians will move in the next few time steps and converts that predicted distribution into a state feature. For each pedestrian $m$, we compute their state visitation distribution $D_{m}^{(t:t+L_m)}$, which describes the likelihood of pedestrian $m$ being in a certain location at a certain time step. The state visitation distributions of all other pedestrians $\bar{D}_{\neg n}$ are then summed together to generate the state feature $f_{n,soc}^{(t)}$.

\noindent\textbf{\textsc{UpdateUtility}}. In order to predict how each pedestrian will move over a sequence of time steps, and to compute how those predictions will affect the predictions of other pedestrian, we need to use a time-varying utility function for each pedestrian $n$,
\begin{align}
U_{n}^{(t)}[\vec{s}_n, \mu_{\neg n}^{(t)} (\vec{s}_{\neg n}) ] \propto \exp \sum_{\vec{x} \in \vec{s}_n} \vec{\theta}^{\top}\vec{f}^{(t)}_n(\vec{x}).
\label{eq:utility-2}
\vspace{-3mm}
\end{align}

Notice that the utility function is also a maximum entropy distribution, where the empirical distribution of all other pedestrians $\mu_{\neg n}^{(t)} (\vec{s}_{\neg n})$ has been incorporated through the feature vector $\vec{f}^{(t)}_n(x)$ (details in Section \ref{sec:features}). The utility function is updated every $\tau$ time steps, the frequency at which each pedestrian makes predictions about the movement of others. A distribution over states sampled from $U_{n}$ in illustrated in Figure \ref{fig:fp-flow} (right).

It is important to make a connection between the utility function $U$ and the empirical distribution $\mu$ at this juncture. In our formulation, $U_m$ is exactly equivalent to $\mu_m (\vec{s}_m)$. In general, $U$ need not be a probability distribution, as it simply describes the value (reward) of one macro-action over another. In contrast, the empirical distribution $\mu$ is a probability distribution by construction and it describes which macro-action the opponent is likely to take. More importantly, the utility function $U$ helps us to understand the dependency of the predicted path of pedestrian $n$ on the predicted path of all other pedestrians $\neg n$. When we use the utility function to forecast the path of a single agent, that prediction influences the predicted path of all other predictions. This interplay between forecasted paths is precisely what we set out to model.

\noindent\textbf{\textsc{TakeMacroAction}}. This function takes the current empirical distribution $\mu_n^{(t)}$ and the prior state visitation distribution $D_n^{(t-1)}$ to compute the 
future state visitation distribution $D_n^{(t:t+L_n)}$. Intuitively, this function forward simulates pedestrian motion $L_n$ steps into the future.
To compute the future state visitation distribution $D_n^{(t:t+L_n)}$, a policy $\pi$ is first derived from the empirical distribution (this process described in Section \ref{sec:ioc}). Using that policy, we iteratively compute how the prior state distribution $D_n^{(t-1)}(\vec{x})$ will change in over the next $L_n$ time steps. 

As a more concrete example, Figure \ref{fig:fp-flow} illustrates the procedure where we employ fictitious play to model the interactions within three pedestrians. The three pedestrians, respectively colored in red, green, and yellow, sequentially make predictions (\ie, fictitious play) of others' macro-actions based on $\mu_n (\vec{s}_{\neg n})$ and then take the macro-action that maximize one's utility function. The forecasted state visitation distribution $\bar{D}_n$ (detailed in Section \ref{sec:features}) of each pedestrian is expressed in the corresponding color and grows incrementally over time.

\subsection{A Decision-Theoretic Pedestrian Model}
\label{sec:ioc}

We now explain how to learn the maximum entropy distribution which is used for both the utility function $U_n$ and independent empirical distributions $\mu_m$. As we have alluded to earlier, the probability of generating a trajectory $\vec{s}$ is modeled to be drawn from a maximum entropy distribution, where the probability is proportional to the exponentiated sum of weighted features encountered over the trajectory,
\begin{align}
    P(\vec{s};\vec{\theta}) = \frac{1}{Z(\vec{\theta})} \exp \sum_{\vec{x} \in \vec{s}} \vec{\theta}^{\top} \vec{f}(\vec{x}),
\end{align}
where $Z$ is the normalization function (or partition function), $\vec{\theta}$ is a vector of parameters and $\vec{f}(\vec{x})$ is a vector of features at state $\vec{x}$. 

In order to learn the parameters $\vec{\theta}$ of this model from a set of demonstrated pedestrian trajectories, we utilize maximum entropy inverse optimal control \cite{ziebart2008maximum}. We first make the assumption that each pedestrian is a rational agent and plans a path according to an underlying Markov Decision Process (MDP). The MDP describing a pedestrian $n$ is defined by an initial state distribution $P_n(\vec{x}_0)$, a transition model $P_n(\vec{x}'|\vec{x},\vec{a})$ and a reward function $R_n^{(t)}(\vec{x})$. Following \cite{kitani2012activity}, the reward function is further defined as a weighted combination of features, $R_n^{(t)}(\vec{x}) = \vec{\theta}^{\top} \vec{f}^{(t)}_n(\vec{x})$. Note however that our reward function $R_n^{(t)}(\vec{x})$ is time indexed, as the feature vector $\vec{f}^{(t)}_n(x)$ will be used to encode information about changes in predicted behaviors of other pedestrians.

To learn the parameters $\vec{\theta}$ using maximum entropy IOC \cite{ziebart2008maximum}, we implement a gradient descent procedure that first computes a policy $\pi(\vec{a}|\vec{s}; \vec{\theta})$ based on the current estimate of $\vec{\theta}$. The we compute the gradient update using difference between the estimated cumulative feature count and empirical cumulative feature count over demonstrated trajectories given that policy. When the features accumulated over trajectories generated by the MDP model converge to values similar to the empirical feature counts of the training data (\ie, likelihood under the maximum entropy distribution is maximized), the algorithm has obtained an optimal set of parameters $\hat{\vec{\theta}}$, which will be used to define the empirical distribution $\mu$.

An optimal policy for the maximum entropy distribution $P(\vec{s};\vec{\theta})$ can be computed as $\pi(\vec{a}|\vec{s}) = \exp\{ Q( \vec{x},\vec{a}) - V(\vec{x})\}$, where the state-action soft value function $Q( \vec{x},\vec{a})$ and state soft-value function $V(\vec{x})$ can be computed by iterating the soft-maximum Bellman update equations: $Q( \vec{x},\vec{a}) = \vec{\theta}^{\top}\vec{f}(\vec{x}) + E_{P(x'|x,a)}[V(\vec{x}')]$ and $V(\vec{x}) = \textrm{softmax}_a Q(\vec{x},\vec{a})$.
We call this procedure \textsc{ComputePolicy} in Algorithm \ref{alg:encode-feature}. Recall that in our scenario, the policy is time-varying since the features of the states change over time. Therefore, the policy for each pedestrian must be recomputed each time features are updated, \ie, every forecasting period $\tau$.

\subsection{Features for Forecasting}
\label{sec:features}

\begin{figure}[tb]
\begin{center}
\scriptsize
\includegraphics[width=0.99\linewidth]{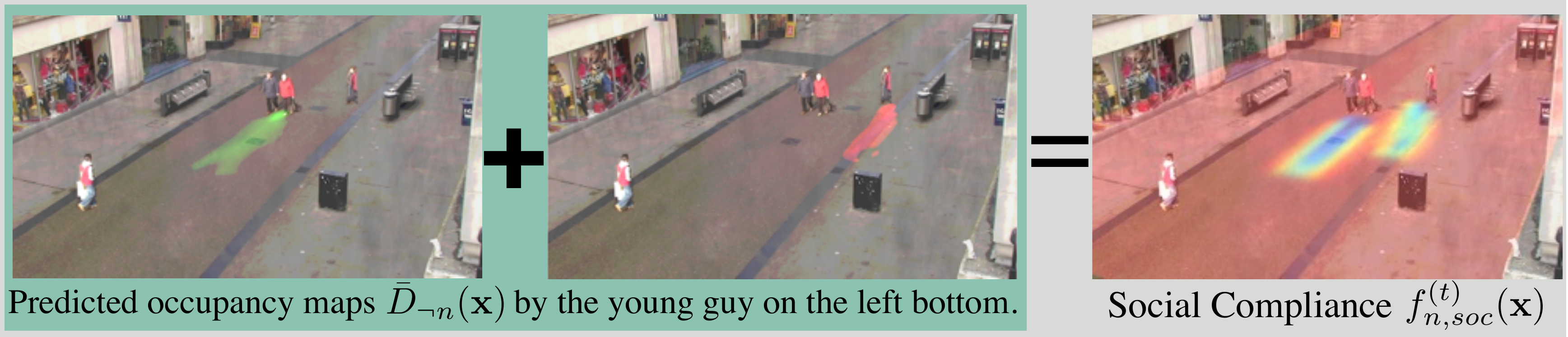}
\end{center}
\vspace{-4mm}
\caption{ The green box demonstrates how a pedestrian forms beliefs about others $\mu_{\neg n}^{(t)}$ and encode such information into social compliance feature $f_{n,soc}^{(t)}(\vec{x})$. {\color{red}Red} indicates high reward and {\color{blue}blue} indicates low reward.}
\label{fig:social-feature-response}
\vspace{-3mm}
\end{figure}

In this section, we first show how the policy can be used to design the time dependent social compliance feature $f_{n, soc}^{(t)}$, which captures the interdependent reasoning process among people by encoding the empirical distribution over trajectories of all other agents $\mu_n^{(t)} (\vec{s}_{\neg n})$ into it. Then we build upon prior work \cite{kitani2012activity} and introduce the semantic scene features which encode the intuition that rational agents will take into account the physical layout of the scene as they plan their future trajectories. Finally, we estimate the initial body orientation of each agent to encourage predictions that are aligned with body directions.

\noindent \textbf{Social Compliance Feature}: Given a policy $\pi(\vec{a}|\vec{x})$, we can generate a state visitation distribution $D_n$ of pedestrian $n$ for trajectories of length $L_n$ by recursively computing:
\begin{align}
    D_n^{(l)}(\vec{x}') =  \sum_{\vec{a},\vec{x}} P(\vec{x}'|\vec{x},\vec{a}) \pi(\vec{a}| \vec{x}) D_n^{(l-1)}(\vec{x}), 
\end{align}
where $D_n^{(0)}(\vec{x})$ needs to be initialized to a distribution over start locations. Since $D_n^{(l)}(\vec{x})$ is defined over the entire state space, it is the same size as the state space. We can sum visitation counts over time, $\bar{D}_n(\vec{x}) = \sum_{l} D_n^{(l)}(\vec{x})$ to generate a cumulative distribution over states. The cumulative state visitation distribution $\bar{D}_n(\vec{x})$ represents the states that are likely to be occupied by pedestrian $n$ when sampling from the empirical distribution $\mu_n (\vec{s}_n)$. By aggregating the cumulative visitation distribution for all pedestrian except $n$, we can obtain a predicted occupancy map of all pedestrians in the environment, $\bar{D}_{\neg n}(\vec{x}) = \sum_{m\neq n} \bar{D}_{m}(\vec{x})$, which we will use to form our social compliance feature $f^{(l)}_{n,soc}$. Formally, this quantity encodes the empirical distribution $\mu_{\neg n} (\vec{s}_{\neg n})$ passed to the utility function (Equation \ref{eq:utility-2}). The process is summarized in Algorithm \ref{alg:encode-feature}. 

More intuitively, this quantity describes a social force field. Based on Helbing and Molonar's model of social forces \cite{helbing1995social} we define several social distance features that places a force field of varying size around the predicted trajectories of all other pedestrians in the environment. In particular, we defined three different sized force-fields, that roughly corresponds to Hall's proxemics zones \cite{hall1966hidden}, to encode a range of physical distances that people may maintain when walking in crowded scenes. Note that the social compliance features $f_{soc}^{(t)}(\vec{x})$ are indexed by time as the predicted paths of other pedestrians changes over time. It is also interesting to note that our feature naturally supports group behavior analysis, even though we do not explicitly model it as in \cite{yamaguchi2011you}. When summing over the cumulative distribution of all other pedestrians, the states nearby groups will have larger visitation counts (\ie. more likely to be occupied), resulting in a more collision prone area. Figure \ref{fig:social-feature-response} shows a situation where the area in front of the couple is a high potential collision area and thus has a lower reward. 

\noindent \textbf{Neighborhood Occupancy:} 
This feature is a measurement of the amount of obstacles in a local neighborhood around a certain state. We calculate the number of pixels labeled as obstacles in a $5 \times 5$ grid and normalize it to provide a soft estimate of whether a state is a obstacle or not. The feature encodes how close pedestrians will walk near static objects in the scene. The neighborhood occupancy feature is denoted as $f_{occ}(\vec{x})$ which is not time varying as we assume the geometry of the scene to be static. 

\noindent \textbf{Distance-to-Goal:} 
The feature $f_{dog}(\vec{x})$ captures a pedestrian's desire to approach his goal quickly by computing the Euclidean distance between a state $\vec{x}$ and the goal $\vec{x}_{g}$. 

\noindent \textbf{Body Orientation}: Since a pedestrian's body orientation is a strong cue of the direction in which she will walk \cite{chamveha2011appearance}, we train a CNN (described in details in Section \ref{sec:appearance}) to predict the initial walking direction of a pedestrian. We use the value of cosine distance minus one over the 8 connected neighbors centered at current pedestrian location. The value is greatest ($0$) in the direction of the predicted velocity direction and is the lowest ($-2$) in the opposite direction. The body orientation feature is denoted as $f_{bod}(\vec{x})$.

\subsection{Walking Characteristics from Appearance}
\label{sec:appearance}

We further enhance the predictive power of our multi-pedestrian framework by allowing the model to maintain individualized walking models for each pedestrian based on appearance. In this section, we focus on visual information which conveys salient cues about how each individual in the scene may walk. For example, when we walk in crowds, the initial body orientation of a person may inform us of which direction that individual might walk. We may also perform high-level visual inference, predicting that an elderly couple might walk slow or a young business man might walk with a brisk pace. We propose using visual classifiers to identify various attributes of a pedestrian, and then map those attributes to walking direction and speed.


To extract attributes from a pedestrian's visual appearance, we make use of a deep learning model. In particular, we employ a network structure similar  to~\cite{krizhevsky2012imagenet}, but modify the top layer to generate three classification outputs: (1) age (old or young), (2) gender (male or female) and (3) body orientation (8 discretized direction). We train all three top layer classifiers jointly, as previous work has shown that multi-task learning helps to constrain the parameter learning \cite{tian2014pedestrian,yuan2012visual}. The predicted body orientation is used to generate the body orientation feature mentioned in Section~\ref{sec:features}, while the output of the age and gender classifiers are used to build individualized pedestrian models. 

To be concrete, we use the soft probabilistic output of the age and gender classifiers to estimate an individualized velocity parameter. For each pedestrian $n$, we compute the individual's velocity $v_n$ as the weighted average over gender and age velocity averages, \ie\ $v_n = \sum_a w_a v^{stats}_a$, where $a \in \{male, female, old, elder\}$ denotes the attributes, $w_a$ denotes the softmax output from deep net, and $v^{stats}_a$ represents the average speed of pedestrians with attribute $a$. The individualized speed $v_n$ is then incorporated into our model by multiplying the forecasting window size $W$, \ie\ $L_n = W \times v_n$. Recall that $L_n$ is the length of macro-actions $s_n$ and $v_n$ denotes speed, $W$ can thus be interpreted as \emph{how many time steps into the future one will predict about others}.  In general, given a fixed $W$, the faster a pedestrian walks, the larger his occupancy map $\bar{D}_n = \sum_{l=t}^{t+W*v_n} D_n^{(l)}$ may be. We note that when speed information is not available, we employ a constant speed $C$ for every pedestrian, \ie, $L_n = W \times C\ \forall n$. We also tried regressing velocity directly from appearance, but in practice deep nets fail to learn discriminative features for direct regression.

\begin{figure}[tb]
\begin{center}
\includegraphics[width=0.96\linewidth, trim = 0 0 0 0, clip]{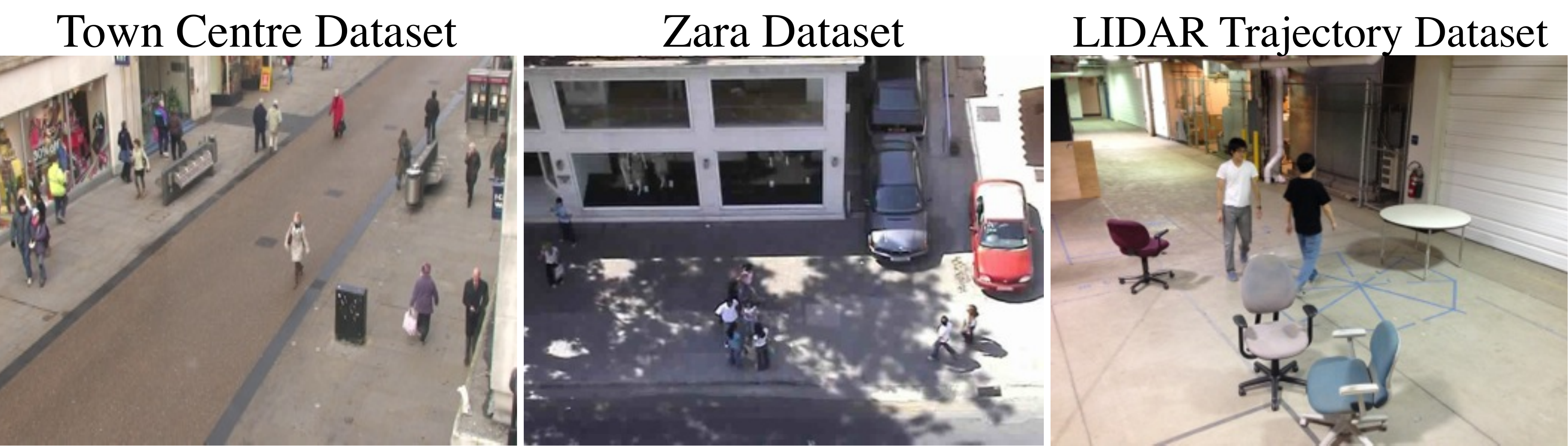}
\end{center}
\vspace{-5mm}
\caption{An overview of the datasets we used in the experiments.}
\vspace{-5mm}
\label{fig:dataset}
\end{figure}

\begin{table*}[!htb]
\begin{center}
\scriptsize
\begin{minipage}[b]{0.49\linewidth}
\centering
\scalebox{0.83}{
\begin{tabular}{| l ||c|c|c|c|c|}
\hline
~NLL~~	&nMDP\cite{kitani2012activity}	&MDPCV	&mTA\cite{pellegrini2009you}&FP &FP + Speed
\\\hline
Zara~\cite{lerner2007crowds}	&46.5396	&46.9549	&43.3834	&{\bf 42.1426} &-
\\\hline
Town Centre \cite{benfold2011stable}	&14.4797	&14.4011	&14.2471	&12.5804 & {\bf 10.892}
\\\hline
LIDAR Trajectory	&92.5579 &93.1747 &91.9748 &{\bf 87.4680} & -
\\\hline
Zara (no-dest) \cite{lerner2007crowds}&98.7343 &97.6634	&92.8271&{\bf 88.5693} &-
\\\hline
Town Centre (no-dest) \cite{benfold2011stable}	&33.8454 & 33.3213	&31.5433	&27.5732 & {\bf 27.2136}\\\hline
LIDAR Trajectory (no-dest)&173.643 &175.384 &169.782 &{\bf 161.338} & -
\\
\hline
\end{tabular}}
\end{minipage}
\begin{minipage}[b]{0.49\linewidth}
\centering
\scalebox{0.83}{
\begin{tabular}{| l ||c|c|c|c|c|}
\hline
SCR~~	&nMDP\cite{kitani2012activity}	&MDPCV	&mTA\cite{pellegrini2009you} &FP &FP + Speed\\\hline
Zara \cite{lerner2007crowds}	&0.144 &0.114 &0.065 &{\bf 0.013}& -\\\hline
Town Centre \cite{benfold2011stable}	&0.215	&0.213	&0.120	&0.052 &{\bf 0.049}\\\hline
LIDAR Trajectory	&0.133    &0.105  &0.056  &{\bf 0.009}& -\\\hline
Zara (no-dest) \cite{lerner2007crowds}	&0.186 &0.175 &0.095 &{\bf 0.021}& -\\\hline
Town Centre (no-dest) \cite{benfold2011stable}	&0.323	&0.281	&0.170	&0.093 &{\bf 0.066}\\\hline
LIDAR Trajectory (no-dest)&0.197    &0.173  &0.082  &{\bf 0.022}& -\\\hline
\end{tabular}}
\end{minipage}
\end{center}
\vspace{-1mm}
\caption{Comparative analysis between different approaches. Smaller values are better.}
\label{tab:fullforecast}
\vspace{-4mm}
\end{table*}

\vspace{-1mm}
\section{Experiments}
\vspace{-2mm}

We analyze our model from various aspects. Following \cite{kitani2012activity}, we first assume the destinations are known to evaluate our forecasting performance in isolation. To validate the effectiveness of our model in the real world, we later perform unconstrained experiments with unknown goals. We evaluate our model on three different pedestrian interaction datasets: the Zara Dataset~\cite{lerner2007crowds}, the Town Centre Dataset~\cite{benfold2011stable}, and the LIDAR Trajectory Dataset. The first two datasets represent real world crowded settings with non-linear trajectories. To show that our model can also work with other modes of trajectory data, we further collected a LIDAR-based Trajectory Dataset, consisting of 20 interactive trajectories. Subjects are initialized at various start locations in a small room ($7m \times 7m$) with a few obstacles. They are then directed to walk towards a goal location without colliding with other pedestrians in the scene. We show samples of each dataset in Figure~\ref{fig:dataset}.

\vspace{-1mm}
\subsection{Metrics and Baselines}
\vspace{-1mm}


\noindent \textbf{Negative Log Loss (NLL).} 
The negative log loss computes the likelihood of drawing the demonstrated trajectory. It is defined as $NLL(\vec{s}) = -\sum_t \log \pi^{(t)}(\vec{a}^{(t)}|\vec{x}^{(t)}),$ where trajectory $\vec{s}$ is a sequence of state-action pairs ($\vec{x}$, $\vec{a}$).

\noindent \textbf{State Collision Rate (SCR).} While the NLL is appropriate for evaluating single agent forecasting results, it does not explicitly penalize colliding predicted paths in the multi-agent forecasting case. To encode the notion of a future collision, we define the State Collision Rate $SCR = \sum_{t} \prod_{n} D^{(t)}_{n}(\vec{x}),$ where $n$ denotes a pedestrian ID, and $D^{(t)}_{n}(\vec{x})$ represents the expected state visit count at state $\vec{x}$ at time $t$, \ie, the probability of being at certain state at a certain time. By taking into account the distribution of multiple pedestrians and taking their union, the resulting state visit count of all agents represents regions of collision.

We compare with the following three baselines:

\noindent\textbf{N-Independent MDP (nMDP)}. This baseline model is the approach of \cite{ziebart2008maximum} applied to images for forecast the trajectory of a single pedestrian \cite{kitani2012activity}. Extending their approach for our multi-agent scenario, we use $N$ instantiations of their MDP model and run them in parallel.

\noindent\textbf{MDP + Constant Velocity (MDPCV)}. The second baseline model is a modification of the Independent MDP model but with a collision region features added to the reward function. By assuming constant velocity, we can compute regions of collision (\ie, the intersection regions of linear motion models) and encode them using the same way as the neighborhood occupancy feature.

\noindent\textbf{mTA}. Based on the work of Pellegrini \etal \cite{pellegrini2009you}, the third approach is a modified Trajectory Avoidance (mTA) model. In~\cite{pellegrini2009you} every agent chooses a velocity that minimizes its energy function at every time step. Formulated as an MDP, this corresponds to a reward function using only a constant feature. As for modeling the social force features (\eg, comfortable distance among agents), we use the social compliance features described in Section \ref{sec:features}. We emphasize here that this baseline model has less information that the original model described in \cite{pellegrini2009you} where every agent knows the positions and velocities of others. This information is not available in our problem setup (\ie, single image input).

\subsection{Multi-Pedestrian Forecasting Performance}
\vspace{-2mm}

\begin{figure*}[tb]
\begin{center}
\includegraphics[width=0.96\linewidth, trim = 0 25 0 0]{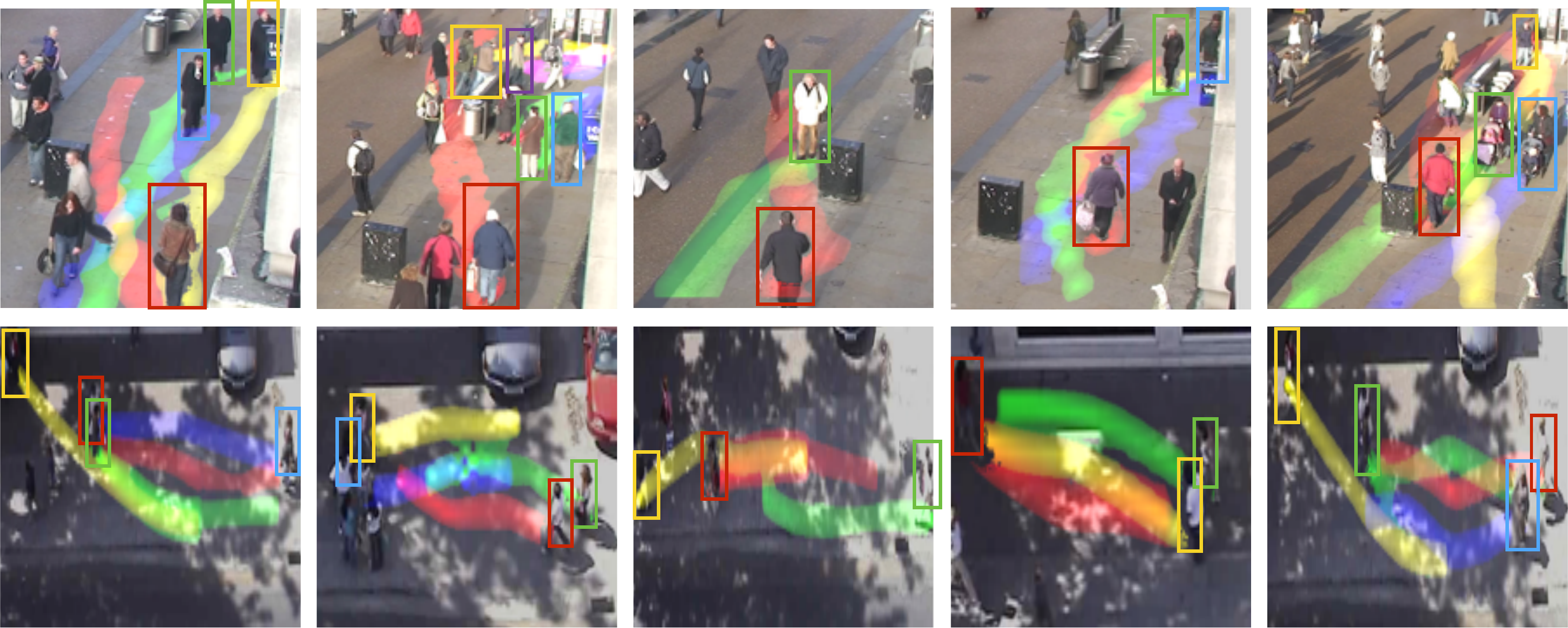}
\end{center}
\caption{Multi-agent forecasting examples of pre-emptive collision avoidance. Each pedestrian is marked with a colored bounding box, with corresponding forecasting distribution in the same color. Note that we consider all pedestrians for quantitative experiments but only visualize forecast distributions for a limited number of pedestrian to improve visualization.}
\label{fig:qualitative}
\end{figure*}

\begin{table}[tb]
\centering
\scalebox{0.9}{
\begin{tabular}{|c||c|c|c|c|}
\hline
&Young &Old &Male & Female\\
\hline
Average Speed (grids/frame)  &1.98 &1.25 &1.78 &1.53\\
\hline
\end{tabular}}
\centering
\scalebox{0.9}{
\begin{tabular}{|c||c|c|c|}
\hline
&Age & Gender & Body Direction\\
\hline
Accuracy &$82.31\%$ &$78.44\%$ &$65.60\%$\\
\hline
\end{tabular}}
\vspace{-0mm}
\caption{Top: average speed of people with different visual attributes. Bottom: accuracy of visual pedestrian classification.}
\label{tab:cnnstats}
\vspace{-3mm}
\end{table}



To properly evaluate our proposed approach, we apply our method only on trajectory sequences that demonstrate strategic reasoning, where multiple pedestrians are actively avoiding each other as they walk. We obtain 16 multi-pedestrian trajectory sequences from each dataset \cite{lerner2007crowds} and \cite{benfold2011stable}\footnote{For more details in how we select the trajectories and the results on the original dataset, please refer to the \href{http://people.csail.mit.edu/weichium/papers/cvpr17/supp.pdf}{supplementary material}.}. We emphasize here that trajectories of single pedestrians simply walking in a straight line are not used, as it is possible to artificially increase  performance by adding more of these `easy' examples. We compute the metrics with 5-fold cross validation.  The forecasting window size is set to $W = 3$ and the forecasting period is $\tau = 1$. The two parameters are found via grid search, and a detailed analysis can be found in the supplementary material. Results are summarized in Table \ref{tab:fullforecast}. We observe that our fictitious play based approach outperforms all three approaches with respect to NLL and SCR. This shows that our iterative predicting and planning process better predicts human interactions and also generates the most collision-free trajectories.



We further incorporate speed information into our model using the method mentioned in Section~\ref{sec:appearance}. We evaluate the effectiveness of speed information on the Town Centre dataset as the resolution of the Zara dataset is too low to extract visual features and the LIDAR-based Trajectory dataset does not provide any visual features. We collected $\approx$16K pedestrian patches from the Town Centre Dataset~\cite{benfold2011stable}, with three labels for each patch, \ie, age, gender, and body orientation.
The images are split into the corresponding 5-fold by pedestrians.
We train a deep classifier using the network structure in Section~\ref{sec:appearance}. The performance is shown in Table \ref{tab:cnnstats}(bottom). We also computed speed statistics from the videos (see Table \ref{tab:cnnstats}(top)). Using the speed statistics and the output of the deep network model, we can compute an individualized model for each pedestrian. As expected, Table \ref{tab:fullforecast} shows that our model performs better after considering a pedestrian's visual appearance and individualize the predictive model. Selected qualitative results of predicted trajectories are shown in Figure \ref{fig:qualitative}.

\noindent\textbf{Relax Destination Constraints.} To show that our model also work in the real work settings where the final destination of a pedestrian is not known and needs to be inferred, we follow \cite{kitani2012activity} to densely generate potential goals on the map and perform the same forecasting experiment. Results, denoted as \emph{no-dest} in Table \ref{tab:fullforecast}, show that our FP based approach, which models the interplay and visual evidence, still consistently outperforms others even without knowing the destinations ahead of time. The absolute performance of all models degrade due to uncertainty about the goal.


\vspace{-2mm}
\subsection{Features Analysis}
\vspace{-2mm}

We further evaluate the effects of the features used in our proposed model. 
The average NLL and SCR for the Town Centre dataset using different features are shown in Table \ref{tab:feature-analysis} (results on other datasets can be found in the supplementary material). We set forecasting window size $W = 3$ and forecasting period $\tau = 1$ as before. The performance of other approaches are also shown for reference. Note that nMDP and MDPCV still consider only scene features and body orientation features even with the social compliance feature checked, since they cannot handle the dynamics in the environment. Our model performs identical to nMDP when considering only semantic scene features and body orientation features. This result is expected as there are no social compliance features to change the cost topology over time. If there is only one agent (which implies there is no social compliant feature), our proposed model reduces to nMDP. We emphasize that with the inclusion of the social compliance feature, our proposed models better explains the interactions between multiple pedestrians. The FP+Speed model attains a NLL of $10.892$ compared to the next best performing model mTA at $14.247$, resulting in a $23.5\%$ improvement in the NLL.

\begin{table}[t]
\centering
\scalebox{0.63}{
\begin{tabular}{|c|cccc||c|c|c|c|c|}
\hline
&$f_{occ}$ & $f_{dog}$ & $f_{bod}$ & $f_{soc}$ &nMDP\cite{kitani2012activity} &MDPCV	&mTA\cite{pellegrini2009you}&FP &FP + Speed  \\
\hline
\hline
\multirow{4}{*}{\rotatebox[origin=c]{90}{NLL}}
&\checkmark & \checkmark & & &14.48 &14.40 &14.48 &14.48 &14.48\\
&\checkmark & \checkmark & \checkmark& &14.45 &14.33 &14.45 &14.45 &14.45\\
&\checkmark & \checkmark & &\checkmark &14.48 &14.40 &14.29 &12.65 &11.01\\
&\checkmark & \checkmark & \checkmark& \checkmark&14.45 &14.33 &14.25 &12.58 &{\bf 10.89}\\
\hline
\end{tabular}
}
\centering
\scalebox{0.63}{
\begin{tabular}{|c|cccc||c|c|c|c|c|}
\hline
&$f_{occ}$ & $f_{dog}$ & $f_{bod}$ & $f_{soc}$ &nMDP\cite{kitani2012activity} &MDPCV	&mTA\cite{pellegrini2009you}	&FP &FP + Speed  \\
\hline
\hline
\multirow{4}{*}{\rotatebox[origin=c]{90}{SCR}}
&\checkmark & \checkmark & & &0.215 &0.183 &0.215 &0.215 &0.215\\
&\checkmark & \checkmark & \checkmark& &0.211 &0.175 &0.211 &0.211 &0.211\\
&\checkmark & \checkmark & &\checkmark &0.215 &0.183 &0.129 &0.046 &0.044 \\
&\checkmark & \checkmark & \checkmark& \checkmark&0.211 &0.175 &0.120 &0.043 &{\bf 0.039}\\
\hline
\end{tabular}
}
\caption{Contribution of each feature to our model.}
\label{tab:feature-analysis}
\vspace{-0.3cm}
\end{table}

\vspace{-1.5mm}
\section{Conclusion}
\vspace{-1.5mm}
We present a novel framework to forecast multi-pedestrian trajectories from a \textit{single image} by directly modeling the interplay between multiple people using concepts from game theory and optimal control. We also develop various predictive models to show how different modes of information help to reason about the future actions of multi-pedestrian scenarios. By building individualized pedestrian models for each person based on his visual appearance, we generate more accurate prediction of multi-pedestrian interactions. We have compared our Fictitious Play based approach  with other state-of-the-art algorithms. Our evaluation on multiple pedestrian interaction datasets has shown that our proposed approach is able to attain more accurate long-term predictions of pedestrian activity.

\section*{Acknowledgement}
This work was supported by JST CREST Grant Number JPMJCR14E1, Japan.

{\small
\bibliographystyle{ieee}
\bibliography{egbib_new}
}

\end{document}